\documentclass[conference]{IEEEtran}
\IEEEoverridecommandlockouts

\usepackage[T1]{fontenc}
\usepackage[utf8]{inputenc}
\usepackage{cite}
\usepackage{amsmath,amssymb,amsfonts}
\usepackage{esvect}
\usepackage{graphicx}
\usepackage{textcomp}
\usepackage{xcolor}
\usepackage{booktabs}
\usepackage{multirow}
\usepackage{bm}

\def\BibTeX{{\rm B\kern-.05em{\sc i\kern-.025em b}\kern-.08em
    T\kern-.1667em\lower.7ex\hbox{E}\kern-.125emX}}

\newcommand{\safeincludegraphics}[2][]{%
  \IfFileExists{#2}{\includegraphics[#1]{#2}}{%
    \fbox{\parbox[c][0.18\textheight][c]{0.94\linewidth}{%
      \centering\textbf{Missing figure file}\\[2pt]\texttt{#2}}}%
  }%
}

\begin{document}

\title{SDDMO-Bench: A Benchmark Suite for Streaming Data-Driven Dynamic Multi-Objective Optimization}

\author{
\IEEEauthorblockN{Wenjie Xiao}
\IEEEauthorblockA{
School of \\
Xiangtan University\\
Xiangtan, China\\
Email: 202421633186@smail.xtu.edu.cn
}
\and
\IEEEauthorblockN{Hui Bai\textsuperscript{*}}
\IEEEauthorblockA{
School of \\
Xiangtan University\\
Xiangtan, China\\
Email: huibai@xtu.edu.cn\\
\textsuperscript{*}Corresponding author
}
\and
\IEEEauthorblockN{Junhao Chen}
\IEEEauthorblockA{
School of \\
Xiangtan University\\
Xiangtan, China\\
Email: 202521633266@smail.xtu.edu.cn
}
}

\maketitle

\begin{abstract}
Streaming data-driven dynamic multi-objective optimization requires algorithms to track time-varying Pareto fronts using only sequential observations under concept drift. However, systematic evaluation remains difficult because real-world problems usually lack ground-truth optima, drift annotations, and controllable conditions, while existing benchmarks provide limited support for standardized comparison. This paper proposes SDDMO-Bench, a benchmark suite that transforms classical dynamic multi-objective test problems into streaming environments by combining intrinsic objective-mapping evolution, controllable distributional drift, and sequential data revelation. By combining five representative time-dependent base functions with six distributional drift patterns, SDDMO-Bench constructs 30 scenarios with diverse levels of non-stationarity, problem complexity, sample-distribution variation, and Pareto-front evolution. 
Experiments with representative evolutionary algorithms demonstrate that SDDMO-Bench provides challenging and discriminative test scenarios, offering a standardized, controllable, and reproducible benchmark for evaluating adaptability, robustness, and Pareto-front tracking in streaming data-driven dynamic multi-objective optimization.
\end{abstract}


\section{Introduction}

Many real-world optimization problems involve unknown or expensive objective functions that cannot be explicitly formulated and must instead be approximated from observed data \cite{jin2005fitness,lim2010surrogate,sun2017surrogate}. Data-driven optimization (DDO) addresses this challenge by learning objective mappings from historical observations rather than relying on analytical models \cite{jin2019ddo,wang2019offline}. However, most existing DDO methods assume that data are available in fixed batches.

In many practical systems, data arrive continuously and sequentially, as in industrial process control, online recommendation, and adaptive traffic management \cite{gama2014survey,wares2019datastream,gaber2005survey}. This observation-driven setting gives rise to streaming data-driven optimization (SDDO), where optimization and model updating must be performed online using newly arriving samples \cite{zhong2024sddobench,zhong2025trace}. Streaming environments are also commonly affected by concept drift, through which the underlying data distribution and objective relationships evolve over time \cite{gama2014survey,tsymbal2004drift,lu2018concept}.

Meanwhile, many real-world problems involve multiple conflicting objectives \cite{emmerich2018tutorial}. Incorporating streaming observations and concept drift into dynamic multi-objective optimization leads to \emph{streaming data-driven dynamic multi-objective optimization problems (SDDMOPs)}, whose goal is to approximate and track time-varying Pareto-optimal fronts using sequentially arriving data \cite{farina2004dynamic,zhong2024sddobench}. Compared with conventional DDO and dynamic multi-objective optimization (DMO), SDDMOPs require algorithms to simultaneously learn unknown objective mappings, adapt to implicit drift, and maintain convergence and diversity under partial and evolving observations.

Existing benchmarks do not fully support the evaluation of these coupled challenges. Classical DMO benchmarks provide controlled time-varying objective functions but generally assume direct access to objective evaluations and predefined environmental changes \cite{farina2004dynamic,huband2006review,jiang2017benchmark}. Recent SDDO benchmarks model sequential observations and concept drift, but mainly focus on single-objective problems \cite{zhong2024sddobench}. Consequently, traditional DMO benchmarks lack streaming and data-driven characteristics, whereas existing streaming benchmarks do not capture multiple objectives or Pareto-front evolution. A standardized and reproducible benchmark suite for SDDMOPs is therefore still needed.

To address this gap, this paper proposes \emph{SDDMO-Bench}, a benchmark suite that transforms classical dynamic multi-objective problems into streaming environments through a time-step-driven data generation mechanism with controllable concept drift. By combining representative base functions with diverse drift patterns, it constructs scenarios with different levels of non-stationarity, problem complexity, and Pareto-front evolution. The main contributions are as follows:
\begin{itemize}
\item We propose SDDMO-Bench, to the best of our knowledge, the first benchmark suite specifically designed for streaming data-driven dynamic multi-objective optimization.
\item We develop a controllable generation mechanism that combines time-dependent multi-objective functions, drift-driven changes in the observed decision distribution, and sequential data revelation. This design separates objective-mapping evolution from distributional drift and supports their joint evaluation under a unified streaming interface.
\item We demonstrate that SDDMO-Bench provides challenging and discriminative test cases for evaluating adaptability, robustness, and Pareto-front tracking under drifting data streams.
\end{itemize}

The remainder of this paper is organized as follows. Section II reviews related work on dynamic multi-objective optimization and streaming data-driven optimization, and discusses the limitations of existing benchmark studies. Section III presents the proposed SDDMO-Bench framework, including its problem formulation, time-dependent base functions, concept-drift patterns, and streaming data generation procedure. Section IV describes the experimental settings and reports the benchmark analyses and comparative results of representative algorithms. Finally, Section V concludes this paper and outlines directions for future research.

\section{Related Work}

\subsection{Streaming Data-Driven Optimization under Dynamic Environments}

Data-driven optimization addresses problems with unknown or expensive objective functions by constructing surrogate models from observed data to guide optimization \cite{jin2019ddo,lim2010surrogate}. In dynamic environments, existing methods mainly improve adaptability through prediction, knowledge transfer, and learning-based strategies. Representative approaches include prediction-and-optimization frameworks \cite{Fu2024TBM}, multi-surrogate learning \cite{Yang2023EvCo}, surrogate transfer \cite{Liu2025STMOEA}, stochastic environmental adaptation \cite{Qiao2025SDRAMTCSO}, and decision-variable analysis \cite{Zhou2020DMOEALEM,Li2020DVCDMOEA}.

Most existing methods, however, assume batch or stage-wise data and direct objective evaluations after environmental changes. In contrast, streaming environments provide only sequential observations under continuously evolving data distributions caused by concept drift \cite{gama2014survey}, requiring surrogate models and optimization strategies to be updated online.

Recent studies have therefore extended data-driven optimization to streaming environments. Zhong et al.~\cite{zhong2024sddobench} first formulated streaming data-driven optimization under concept drift, followed by drift-aware evolutionary optimization \cite{Zhong2025DASE} and surrogate-transfer-based streaming dynamic multi-objective optimization \cite{Liu2025STMOEA}. These studies establish the algorithmic foundation for streaming data-driven optimization while revealing the need for systematic performance evaluation under streaming dynamic environments.

\subsection{Benchmark Problems for Streaming Data-Driven Dynamic Optimization}

Benchmark problems provide standardized and controllable environments for evaluating optimization algorithms. For dynamic optimization, benchmark design has evolved from early time-varying variants of classical functions \cite{goldberg1987dynamic,dasgupta1992dynamic} to more general benchmark generators, such as the Moving Peaks Benchmark (MPB) \cite{branke1999mpb}, the Generalized Dynamic Benchmark Generator (GDBG) \cite{li2008gdbg}, and the Generalized Moving Peaks Benchmark (GMPB) \cite{li2013gmpb}. However, these benchmarks assume explicit objective functions and optimizer-controlled evaluations, making them unsuitable for streaming data-driven optimization where optimization relies solely on sequential observations.

To model streaming optimization scenarios, Zhong et al.~\cite{zhong2024sddobench} proposed SDDO-Bench by integrating sequential data generation with controllable concept drift. Although it provides a standardized benchmark for streaming data-driven optimization, it is limited to single-objective problems and cannot model Pareto-front evolution in multi-objective optimization.

For streaming data-driven dynamic multi-objective optimization, existing studies have mainly generated streaming decision--objective pairs from dynamic benchmark functions for algorithm validation \cite{Liu2025STMOEA}. Such problem constructions are tailored to individual studies rather than reusable benchmark suites, and they do not provide unified mechanisms for configurable concept drift, sequential data revelation, or standardized evaluation protocols.

Consequently, existing benchmarks either focus on single-objective streaming optimization or adopt problem-specific experimental settings for multi-objective optimization. To the best of our knowledge, no benchmark currently provides a standardized, controllable, and reusable evaluation framework that jointly models sequential data arrival, concept drift, and Pareto-front evolution for streaming data-driven dynamic multi-objective optimization. This limitation motivates the development of the proposed \emph{SDDMO-Bench}.

\section{SDDMO-Bench}
\label{sec:benchmark}

SDDMO-Bench integrates time-dependent multi-objective base functions, controllable distributional drift, drift-driven sampling transformations, and sequential data revelation. The base function $\mathbf{F}_\tau$ governs the evolving objective mapping and Pareto structures, whereas $\Phi(\cdot,\delta_\tau)$ changes only the distribution and coverage of the observed decision vectors. This design separates objective-mapping evolution from distributional drift while allowing them to occur simultaneously.

The benchmark combines five dynamic base functions (F1--F5) with six drift patterns (D1--D6), producing 30 instances. A fixed set of anchor samples is generated within the feasible domain. In each environment, these samples are transformed according to the current drift signal and evaluated using the corresponding time-dependent base function, after which only a subset of the resulting decision--objective pairs is sequentially revealed to the optimizer.

\subsection{Problem Formulation of SDDMOPs}
\label{subsec:problem_formulation}

An SDDMOP is defined over a sequence of discrete environments
$\tau\in\{0,1,\ldots,T-1\}$, where $T$ is the total number of environmental states. In environment $\tau$, the underlying multi-objective problem is formulated as
\begin{equation}
\min_{\mathbf{x}\in\Omega}\mathbf{F}_{\tau}(\mathbf{x})
=
\left(
f_{1,\tau}(\mathbf{x}),\ldots,f_{M,\tau}(\mathbf{x})
\right).
\label{eq:sddmop}
\end{equation}
where $\mathbf{x}\in\Omega\subset\mathbb{R}^{D}$ is a $D$-dimensional decision vector, $M$ is the number of objectives, and $F_t:\Omega\rightarrow\mathbb{R}^{M}$ denotes the time-varying objective mapping.

Unlike conventional query-based dynamic optimization, $F_t$ is not directly accessible and the optimizer cannot actively evaluate arbitrary candidate solutions. Instead, it receives sequential observations in each environment:
\begin{equation}
\mathcal{S}_{\tau}
=
\left\{
(\mathbf{x}_{i,\tau},\mathbf{y}_{i,\tau})
\right\}_{i=1}^{n_{\tau}},
\qquad
\mathbf{y}_{i,\tau}
=
\mathbf{F}_{\tau}(\mathbf{x}_{i,\tau}).
\label{eq:stream}
\end{equation}
where $S_t$ contains the samples observed in environment $t$ and $n_t$ is the number of received samples. The equality $\mathbf{y}_{i,t}=F_t(\mathbf{x}_{i,t})$ describes only the latent data-generating relationship; the observations arrive sequentially and are partially available at any given time.

The optimizer must therefore infer the evolving objective mappings from the available stream and approximate the sequence of time-varying Pareto fronts $\{PF_t\}_{t=1}^{T}$. Compared with conventional DMOPs and actively queried black-box optimization, this observation-driven setting introduces additional challenges arising from partial observability, sequential and potentially non-i.i.d.\ data arrival, model uncertainty, and concept drift.

\subsection{Multi-Objective Base Functions}
\label{subsec:base_functions}

The five base functions F1--F5 correspond to FDA4, FDA5, DF10, DF11, and DF12, respectively. They are selected from the FDA~\cite{farina2004dynamic} and DF~\cite{Jiang2018CEC} benchmark families because they provide widely used, analytically tractable, and structurally diverse dynamic multi-objective problems.

FDA4 and FDA5 mainly describe smooth temporal evolution of Pareto-optimal sets and fronts. DF10, DF11, and DF12 introduce more complex characteristics, including nonlinear temporal variation, changing landscape difficulty, moving decision regions, and time-varying variable interactions. These differences enable the benchmark to evaluate algorithm behavior under heterogeneous dynamic structures.

The base functions provide intrinsic time dependence in the objective landscape. SDDMO-Bench further introduces drift-driven sampling transformations in the decision space. These transformations alter the distribution and coverage of the decision vectors observed by the optimizer, whereas the time-dependent base functions determine the evolution of the objective mapping and Pareto structures. The two sources of non-stationarity therefore operate jointly but play distinct roles in the benchmark. Although FDA/DF-based functions cannot reproduce all irregularities, constraints, and uncertainties of real-world applications, they provide controlled and reproducible structures for isolating the effects of streaming observations and concept drift.

\subsection{Drift Signal Modeling}
\label{subsec:drift_modeling}

SDDMO-Bench employs a scalar drift signal $\delta_\tau=d(\tau;P,\epsilon,k,T)$ to control additional distributional changes in the observed decision vectors, where $P$, $\epsilon$, $k$, and $T$ denote the drift period, noise magnitude, number of abrupt changes, and total number of environments, respectively. Six representative patterns are considered. D1 disables the additional distributional transformation and retains only the intrinsic dynamics of the selected base function. D2 and D3 represent irregular and periodic abrupt changes, respectively; D4 models smooth periodic incremental drift; D5 introduces stochastic perturbations into the incremental pattern; and D6 combines high-frequency oscillation, stochastic noise, and random jumps to represent highly irregular environments.

These patterns cover different combinations of abruptness, periodicity, continuity, and stochasticity rather than monotonically increasing levels of difficulty. The resulting signal $\delta_{\tau}$ controls the transformation operator $\boldsymbol{\Phi}(\cdot,\delta_{\tau})$, while the selected FDA/DF base function determines the intrinsic evolution of the objective landscape. Their combination produces coupled temporal changes: $\delta_\tau$ controls the distribution and coverage of the observed decision vectors through $\Phi(\cdot,\delta_\tau)$, while the selected FDA/DF function controls the evolving objective mapping and Pareto structures. Detailed mathematical definitions and parameter distributions of D1--D6 are provided in the \emph{Drift Signal Modeling} subsection of the Supplementary Material. Supplementary Figs. S1 and S2 visualize the six distributional drift signals and the corresponding movements of representative transformed decision samples, respectively.

\subsection{Drift-Driven Sampling Transformation}

SDDMO-Bench changes the observed decision distribution through range-scaled translation, normalized pairwise rotation, and boundary projection. The operator $\Phi(\cdot,\delta_\tau):\Omega\rightarrow\Omega$ is a sampling transformation and does not redefine $\mathbf{F}_\tau$. Let $\boldsymbol{\Delta}=\mathbf{u}-\boldsymbol{\ell}$ and let $\rho\geq 0$ be a scalar severity coefficient. After translation, $\widetilde{\mathbf{x}}_i=\mathbf{x}_i+\delta_\tau\rho\boldsymbol{\Delta}$ is normalized as $\mathbf{q}_i=2\operatorname{diag}(\boldsymbol{\Delta})^{-1}(\widetilde{\mathbf{x}}_i-\boldsymbol{\ell})-\mathbf{1}$. With $z=\operatorname{clip}(\delta_\tau,-1,1)$, $\theta=4\arcsin(z^2)$, and pairwise rotation matrix $\mathbf{R}(\theta)$, the transformed sample is
\begin{equation}
\mathbf{x}_i^{(\tau)}=\Phi(\mathbf{x}_i,\delta_\tau)=\Pi_\Omega\left(\boldsymbol{\ell}+\frac{1}{2}\operatorname{diag}(\boldsymbol{\Delta})\left(\mathbf{R}(\theta)^\top\mathbf{q}_i+\mathbf{1}\right)\right),
\label{eq:drift_transformation}
\end{equation}
where $\Pi_\Omega$ denotes component-wise boundary projection. Normalization prevents variables with larger numerical ranges from dominating the rotation.

\subsection{Streaming Data Generation Mechanism}

A fixed anchor set $\{\mathbf{x}_i\}_{i=1}^{N}$ is generated once within $\Omega$. At environment $\tau$, each anchor sample is transformed and evaluated as
\begin{equation}
\mathbf{x}_i^{(\tau)}=\Phi(\mathbf{x}_i,\delta_\tau),\qquad \mathbf{y}_i^{(\tau)}=\mathbf{F}_\tau\left(\mathbf{x}_i^{(\tau)}\right).
\label{eq:stream_generation}
\end{equation}
The latent dataset is $\mathcal{D}_\tau=\{(\mathbf{x}_i^{(\tau)},\mathbf{y}_i^{(\tau)})\}_{i=1}^{N}$, from which only $n_\tau$ pairs are revealed as $\mathcal{S}_\tau$. Thus, $\Phi$ changes the marginal distribution of the observed decisions, whereas $\mathbf{F}_\tau$ determines the conditional objective mapping and Pareto structures. Consequently, the reference Pareto sets and fronts are those of $\mathbf{F}_\tau$ over $\Omega$ and are not transformed by $\Phi$. The complete drift-pattern definitions and additional temporal tracking visualizations are provided in the Supplementary Material.

\begin{figure*}[!t]
\centering
\includegraphics[width=0.95\textwidth]{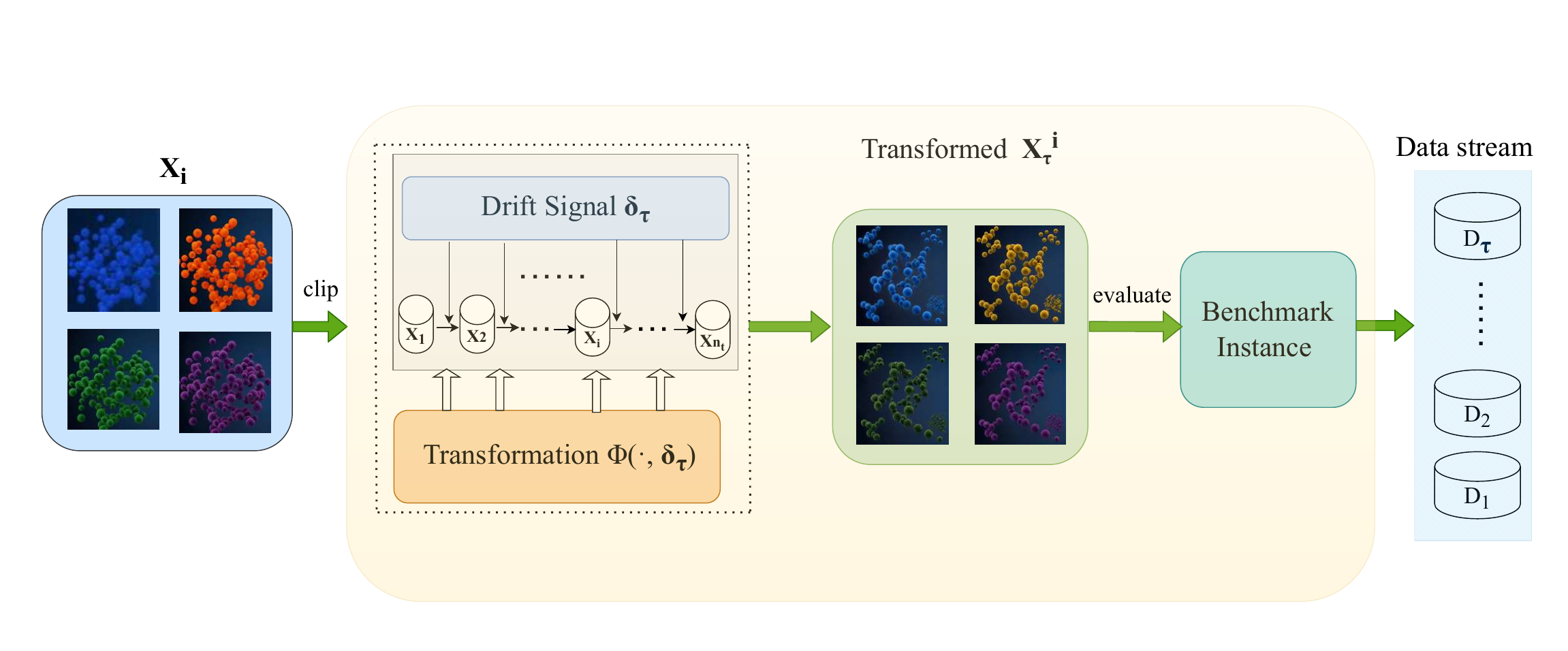}
\caption{Illustration of the streaming data generation mechanism in SDDMO-Bench.}
\label{fig:data_generation}
\end{figure*}

\section{Experimental Studies}

This section evaluates whether SDDMO-Bench provides diverse, challenging, and discriminative scenarios for streaming data-driven dynamic multi-objective optimization. Six representative dynamic and data-driven evolutionary algorithms are employed as analytical tools. The experiments examine: (i) performance variations across the generated scenarios, (ii) the influence of different drift patterns, and (iii) whether the benchmark can reveal heterogeneous algorithmic responses under streaming concept drift.

\subsection{Experimental Setup}
\label{sec:Experimental Setup}

\subsubsection{Benchmark Configuration}

The evaluation examines the diversity, difficulty, and discriminative capability of SDDMO-Bench under different drift patterns. Five classical DMOPs, namely FDA4, FDA5, DF10, DF11, and DF12, are selected as base functions. These problems exhibit different characteristics in Pareto-set evolution, Pareto-front geometry, and variable interactions, as described in Section~\ref{subsec:base_functions}.

\subsubsection{Parameter Settings}

All experiments use unified settings. The decision dimension and number of objectives are set to $D=12$ and $M=3$, respectively. Each run contains $T=60$ environments. The drift period is $P=20$, the noise coefficient is $\epsilon=0.05$, and the drift severity coefficient is $\rho=0.4$. For D2, the number of abrupt changes is set to $k=5$, while the jump probability in D6 is set to $p_J=0.1$. Following the standard FDA/DF settings, the environmental change severity is set to 10, and the change frequency is consistent with $P$.

The population size is 100, and each algorithm evolves for 250 generations in each environment. Every algorithm is independently executed 20 times on each benchmark scenario. The mean inverted generational distance (MIGD), computed by averaging the environment-wise IGD values over all $T=60$ environments, is used to evaluate convergence, solution-set quality, and temporal tracking performance. Its complete definition is provided in Supplementary Section~S2.

\begin{table*}[!t]
\centering
\tiny
\setlength{\tabcolsep}{2.2pt}
\renewcommand{\arraystretch}{1.08}
\caption{MIGD results obtained by six representative algorithms on 30 SDDMO-Bench scenarios. Each entry reports the mean and standard deviation over independent runs. The average ranks are included only as an overall summary of behavior across the benchmark scenarios.}
\label{tab:migd_results}
\resizebox{\textwidth}{!}{%
\begin{tabular}{|l|l|c|c|c|c|c|c|}
\hline
Base Func. & Drift & DNSGA-II & MOEA/D-KF & MSAS-DMOA & PBDMO & ST-MOEA & TBM \\
\hline

\multirow{6}{*}{FDA4}
& D1 & $\mathbf{0.0698\pm0.0007}$ & $0.0767\pm0.0010$ & $0.0725\pm0.0010$ & $0.0700\pm0.0004$ & $0.2334\pm0.0045$ & $0.2350\pm0.0011$ \\ \cline{2-8}
& D2 & $0.4347\pm0.0301$ & $0.3930\pm0.0028$ & $0.3927\pm0.0042$ & $\mathbf{0.3645\pm0.0046}$ & $0.4555\pm0.0070$ & $0.4389\pm0.0044$ \\ \cline{2-8}
& D3 & $0.4056\pm0.0096$ & $0.4342\pm0.0091$ & $0.3851\pm0.0033$ & $\mathbf{0.3755\pm0.0020}$ & $0.4640\pm0.0037$ & $0.4362\pm0.0019$ \\ \cline{2-8}
& D4 & $0.5482\pm0.0097$ & $0.5990\pm0.0093$ & $\mathbf{0.5213\pm0.0034}$ & $0.5331\pm0.0054$ & $0.5988\pm0.0081$ & $0.5819\pm0.0049$ \\ \cline{2-8}
& D5 & $0.5185\pm0.0079$ & $0.5753\pm0.0090$ & $\mathbf{0.4909\pm0.0071}$ & $0.5029\pm0.0073$ & $0.5993\pm0.0198$ & $0.5710\pm0.0086$ \\ \cline{2-8}
& D6 & $0.4885\pm0.0170$ & $0.6147\pm0.0303$ & $\mathbf{0.4642\pm0.0147}$ & $0.4680\pm0.0145$ & $0.6167\pm0.0158$ & $0.5690\pm0.0248$ \\
\hline

\multirow{6}{*}{FDA5}
& D1 & $\mathbf{0.0695\pm0.0008}$ & $0.0762\pm0.0009$ & $0.0718\pm0.0006$ & $0.0697\pm0.0004$ & $0.2332\pm0.0076$ & $0.2338\pm0.0021$ \\ \cline{2-8}
& D2 & $0.2647\pm0.0058$ & $0.2905\pm0.0162$ & $\mathbf{0.2575\pm0.0009}$ & $0.2585\pm0.0013$ & $0.4921\pm0.0078$ & $0.4754\pm0.0059$ \\ \cline{2-8}
& D3 & $0.3481\pm0.0050$ & $0.3984\pm0.0295$ & $0.3359\pm0.0007$ & $\mathbf{0.3356\pm0.0006}$ & $0.5560\pm0.0077$ & $0.5150\pm0.0030$ \\ \cline{2-8}
& D4 & $0.4314\pm0.0079$ & $0.5174\pm0.0125$ & $\mathbf{0.4025\pm0.0024}$ & $0.4297\pm0.0102$ & $0.6727\pm0.0107$ & $0.6590\pm0.0085$ \\ \cline{2-8}
& D5 & $0.4451\pm0.0118$ & $0.4819\pm0.0162$ & $\mathbf{0.3426\pm0.0031}$ & $0.3601\pm0.0113$ & $0.6651\pm0.0126$ & $0.6433\pm0.0170$ \\ \cline{2-8}
& D6 & $0.4230\pm0.0205$ & $0.2198\pm0.0174$ & $0.1847\pm0.0060$ & $\mathbf{0.1706\pm0.0051}$ & $0.6598\pm0.0209$ & $0.6228\pm0.0258$ \\
\hline

\multirow{6}{*}{DF10}
& D1 & $0.1375\pm0.0007$ & $0.1377\pm0.0008$ & $0.1403\pm0.0004$ & $\mathbf{0.1368\pm0.0004}$ & $0.3181\pm0.0087$ & $0.3031\pm0.0010$ \\ \cline{2-8}
& D2 & $0.2445\pm0.0016$ & $0.2501\pm0.0059$ & $0.2424\pm0.0006$ & $\mathbf{0.2418\pm0.0004}$ & $0.4159\pm0.0052$ & $0.3874\pm0.0037$ \\ \cline{2-8}
& D3 & $0.3630\pm0.0034$ & $0.3773\pm0.0171$ & $0.3549\pm0.0009$ & $\mathbf{0.3537\pm0.0008}$ & $0.5348\pm0.0063$ & $0.5031\pm0.0040$ \\ \cline{2-8}
& D4 & $0.4043\pm0.0031$ & $0.4532\pm0.0093$ & $\mathbf{0.3979\pm0.0012}$ & $0.3988\pm0.0010$ & $0.5571\pm0.0063$ & $0.5549\pm0.0074$ \\ \cline{2-8}
& D5 & $0.4009\pm0.0072$ & $0.4278\pm0.0163$ & $\mathbf{0.3421\pm0.0013}$ & $0.3445\pm0.0017$ & $0.5704\pm0.0070$ & $0.5531\pm0.0099$ \\ \cline{2-8}
& D6 & $0.3858\pm0.0110$ & $0.2142\pm0.0189$ & $0.2011\pm0.0027$ & $\mathbf{0.1822\pm0.0045}$ & $0.5975\pm0.0154$ & $0.5458\pm0.0173$ \\
\hline

\multirow{6}{*}{DF11}
& D1 & $\mathbf{0.0499\pm0.0013}$ & $0.0952\pm0.0020$ & $0.0833\pm0.0028$ & $0.0792\pm0.0027$ & $0.2736\pm0.0147$ & $0.3341\pm0.0020$ \\ \cline{2-8}
& D2 & $0.2093\pm0.0204$ & $0.2693\pm0.0339$ & $\mathbf{0.1728\pm0.0059}$ & $0.1801\pm0.0061$ & $0.4591\pm0.0117$ & $0.4977\pm0.0079$ \\ \cline{2-8}
& D3 & $0.2175\pm0.0145$ & $0.3050\pm0.0693$ & $\mathbf{0.1900\pm0.0042}$ & $0.2108\pm0.0084$ & $0.5235\pm0.0118$ & $0.5630\pm0.0061$ \\ \cline{2-8}
& D4 & $0.3880\pm0.0153$ & $0.5534\pm0.0315$ & $\mathbf{0.2973\pm0.0034}$ & $0.3413\pm0.0114$ & $0.6391\pm0.0118$ & $0.6771\pm0.0067$ \\ \cline{2-8}
& D5 & $0.4233\pm0.0110$ & $0.4225\pm0.0233$ & $\mathbf{0.2109\pm0.0035}$ & $0.2309\pm0.0127$ & $0.6415\pm0.0101$ & $0.6455\pm0.0164$ \\ \cline{2-8}
& D6 & $0.3885\pm0.0226$ & $0.1517\pm0.0068$ & $0.1105\pm0.0022$ & $\mathbf{0.1019\pm0.0029}$ & $0.6649\pm0.0265$ & $0.6081\pm0.0244$ \\
\hline

\multirow{6}{*}{DF12}
& D1 & $0.0759\pm0.0008$ & $0.0786\pm0.0006$ & $0.0956\pm0.0021$ & $\mathbf{0.0758\pm0.0006}$ & $0.3357\pm0.0075$ & $0.2991\pm0.0029$ \\ \cline{2-8}
& D2 & $0.2089\pm0.0063$ & $0.2517\pm0.0135$ & $0.2040\pm0.0022$ & $\mathbf{0.2039\pm0.0037}$ & $0.5075\pm0.0128$ & $0.5219\pm0.0066$ \\ \cline{2-8}
& D3 & $0.3080\pm0.0055$ & $0.4178\pm0.0363$ & $\mathbf{0.2960\pm0.0013}$ & $0.3012\pm0.0015$ & $0.6282\pm0.0093$ & $0.5633\pm0.0032$ \\ \cline{2-8}
& D4 & $0.4137\pm0.0141$ & $0.5072\pm0.0245$ & $\mathbf{0.3575\pm0.0057}$ & $0.3737\pm0.0114$ & $0.7138\pm0.0141$ & $0.7187\pm0.0077$ \\ \cline{2-8}
& D5 & $0.4308\pm0.0101$ & $0.4373\pm0.0200$ & $\mathbf{0.3019\pm0.0026}$ & $0.3215\pm0.0152$ & $0.7131\pm0.0115$ & $0.6918\pm0.0130$ \\ \cline{2-8}
& D6 & $0.4629\pm0.0127$ & $0.2981\pm0.0164$ & $0.2396\pm0.0060$ & $\mathbf{0.2005\pm0.0052}$ & $0.7173\pm0.0204$ & $0.6711\pm0.0249$ \\
\hline

\multicolumn{2}{|c|}{Avg. Rank}
& $2.933$
& $3.867$
& $1.733$
& $\mathbf{1.600}$
& $5.667$
& $5.200$ \\
\hline
\end{tabular}%
}
\end{table*}

\subsection{Selected Algorithms}

Six representative algorithms are selected as analytical tools: DNSGA-II~\cite{deb2007dynamic}, MOEA/D-KF~\cite{Muruganantham2016KF}, MSAS-DMOA~\cite{Li2024MSAS}, PBDMO~\cite{Zhang2020PBDMO}, ST-MOEA~\cite{Liu2025STMOEA}, and TBM~\cite{Fu2024TBM}.

DNSGA-II introduces population diversity after environmental changes. MOEA/D-KF combines decomposition-based optimization with Kalman-filter prediction, while MSAS-DMOA employs multi-strategy adaptive selection and transfer learning. PBDMO uses prediction-based response strategies to generate populations for new environments. ST-MOEA transfers surrogate knowledge across environments, whereas TBM integrates data-driven prediction with multi-objective optimization.

These algorithms represent diversity-, prediction-, transfer-, and surrogate-assisted response mechanisms. They are used to examine whether SDDMO-Bench can generate scenarios with different levels of difficulty and reveal heterogeneous algorithmic responses, rather than to establish a definitive ranking among the compared methods.

\subsection{Results and Analysis}

Table~\ref{tab:migd_results} reports the MIGD results of six representative algorithms on 30 SDDMO-Bench scenarios. The experiments assess the empirical reliability, validity, applicability, and discriminative capability of the benchmark rather than establish a definitive ranking of the compared algorithms.

Because MIGD averages performance over all environments, it cannot fully reveal algorithm behavior in individual environments. Supplementary Section~S3 therefore presents representative temporal approximation fronts on F3 (DF10) under D1--D6. Supplementary Figs.~S3--S8 show the results of DNSGA-II, MOEA/D-KF, MSAS-DMOA, PBDMO, ST-MOEA, and TBM at $\tau=0$, $\tau=30$, and $\tau=59$, respectively. The observed differences in convergence, coverage, dispersion, and post-change recovery complement the aggregate MIGD results and illustrate heterogeneous tracking behavior under different drift patterns.

\subsubsection{Reliability and Validity of Benchmark Configurations}

The repeated evaluations exhibit generally stable outcomes. Among the 180 algorithm--scenario combinations, approximately 89\% of the standard deviations are below 0.02 and 97\% are below 0.03. In comparison, the scenario-level mean MIGD ranges from approximately 0.126 on FDA5-D1 to 0.564 on FDA4-D4. The cross-scenario differences are therefore substantially larger than most run-to-run variations, supporting the empirical repeatability of the evaluation.

The drift patterns also produce systematic and measurable effects. Averaged over all base functions and algorithms, the MIGD values under D1--D6 are approximately 0.152, 0.333, 0.400, 0.508, 0.477, and 0.408, respectively. D1, which disables the additional decision-space transformation, provides the lowest aggregate error, whereas D2--D6 introduce substantially higher tracking errors. The mean standard deviation also increases from approximately 0.0024 under D1 to 0.0151 under D6, indicating greater adaptation variability under the hybrid drift pattern.

The difficulty is not monotonic across D1--D6: D4 yields the highest aggregate MIGD, while D6 is not uniformly the most difficult. This is expected because the six patterns represent qualitatively different temporal dynamics rather than ordered severity levels. Their effects depend on the interaction among the drift pattern, transformed sample distribution, base-function structure, and algorithmic response mechanism. Thus, SDDMO-Bench generates controllable but nontrivial environmental variations.

\subsubsection{Applicability and Discriminative Capability}

The overall mean MIGD values for FDA4, FDA5, DF10, DF11, and DF12 are approximately 0.433, 0.378, 0.360, 0.339, and 0.387, respectively. These differences indicate that the drift-driven mechanism preserves heterogeneous Pareto-front geometries, Pareto-set dynamics, and variable interactions rather than overwhelming the intrinsic characteristics of the base functions.

The six algorithms cover diversity-based, state-prediction, population-prediction, transfer-learning, and surrogate-assisted response mechanisms, and all are evaluated through the same streaming interface, environmental schedule, computational budget, and temporal metrics. Within individual scenarios, the difference between the largest and smallest MIGD values ranges from approximately 0.078 to 0.563, while the average ranks range from 1.600 to 5.667. Moreover, different algorithms obtain the lowest MIGD under different conditions. These results show that SDDMO-Bench supports multiple algorithmic paradigms and distinguishes their sensitivities to different problem structures and drift patterns without consistently favoring a single response mechanism.

Overall, the small run-to-run deviations and substantial variations across drift patterns, base functions, and algorithms support the empirical reliability, validity, applicability, and discriminative capability of SDDMO-Bench.

\section{Conclusion}

This paper proposed SDDMO-Bench, a benchmark suite for streaming data-driven dynamic multi-objective optimization. By integrating dynamic base functions, controllable drift-driven transformations, and sequential data revelation, the benchmark constructs diverse non-stationary environments that capture both evolving objective landscapes and streaming observations. Experimental results demonstrate that SDDMO-Bench produces reliable, challenging, and discriminative scenarios across different problem structures, drift patterns, and algorithmic response mechanisms.

The current benchmark mainly relies on classical FDA/DF base functions, which provide controllable dynamics and tractable reference fronts but cannot fully represent the constraints, uncertainties, irregular structures, and domain-specific characteristics of real-world applications. Future work will incorporate real-world-inspired, constrained, and simulation-based problems, together with richer drift patterns and higher-dimensional objective spaces.


\begin{thebibliography}{46}

\bibitem{jin2019ddo}
Jin, Y., Wang, H., Chugh, T., Guo, D., Miettinen, K.:
Data-driven evolutionary optimization: An overview and case studies.
\textit{IEEE Transactions on Evolutionary Computation} \textbf{23}(3), 442--458 (2019)

\bibitem{jin2005fitness}
Jin, Y.:
A comprehensive survey of fitness approximation in evolutionary computation.
\textit{Soft Computing} \textbf{9}(1), 3--12 (2005)

\bibitem{lim2010surrogate}
Lim, D.H., Jin, Y., Ong, Y.S., Sendhoff, B.:
Generalizing surrogate-assisted evolutionary computation.
\textit{IEEE Transactions on Evolutionary Computation} \textbf{14}(3), 329--355 (2010)

\bibitem{wang2019offline}
Wang, H., Jin, Y.:
Offline data-driven evolutionary optimization using selective surrogate ensembles.
\textit{IEEE Transactions on Evolutionary Computation} \textbf{23}(2), 203--216 (2019)

\bibitem{sun2017surrogate}
Sun, C., Jin, Y., Cheng, R., Ding, J., Zeng, J.:
Surrogate-assisted cooperative swarm optimization of high-dimensional expensive problems.
\textit{IEEE Transactions on Evolutionary Computation} \textbf{21}(4), 644--660 (2017)

\bibitem{gama2014survey}
Gama, J., Zliobaite, I., Bifet, A., Pechenizkiy, M., Bouchachia, A.:
A survey on concept drift adaptation.
\textit{ACM Computing Surveys} \textbf{46}(4), 44 (2014)

\bibitem{wares2019datastream}
Wares, S., Isaacs, J., Elyan, E.:
Data stream mining: Methods and challenges for handling evolving data streams.
\textit{SN Applied Sciences} \textbf{1}, 1415 (2019)

\bibitem{gaber2005survey}
Gaber, M.M., Zaslavsky, A., Krishnaswamy, S.:
Mining data streams: A review.
\textit{ACM SIGMOD Record} \textbf{34}(2), 18--26 (2005)

\bibitem{zhong2024sddobench}
Zhong, Y., Wang, X., Sun, Y., Gong, Y.-J.:
SDDObench: A benchmark for streaming data-driven optimization with concept drift.
In: \textit{Proceedings of the Genetic and Evolutionary Computation Conference (GECCO)}, pp.~1--9 (2024)

\bibitem{zhong2025trace}
Zhong, Y., Huang, T., Xiao, X., Gong, Y.-J.:
TRACE: A generalizable drift detector for streaming data-driven optimization.
\textit{arXiv preprint arXiv:2512.07082} (2025)

\bibitem{emmerich2018tutorial}
Emmerich, M., Deutz, A.:
A tutorial on multiobjective optimization: fundamentals and evolutionary methods.
\textit{Natural Computing} \textbf{17}(3), 585--609 (2018)

\bibitem{tsymbal2004drift}
Tsymbal, A.:
The problem of concept drift: definitions and related work.
Computer Science Department, Trinity College Dublin (2004)

\bibitem{lu2018concept}
Lu, J., Liu, A., Dong, F., Gu, F., Gama, J., Zhang, G.:
Learning under concept drift: A review.
\textit{IEEE Transactions on Knowledge and Data Engineering} \textbf{31}(12), 2346--2363 (2018)


\bibitem{farina2004dynamic}
Farina, M., Deb, K., Amato, P.:
Dynamic multiobjective optimization problems: Test cases, approximations, and applications.
\textit{IEEE Transactions on Evolutionary Computation} \textbf{8}(5), 425--442 (2004)

\bibitem{huband2006review}
Huband, S., Hingston, P., Barone, L., While, L.:
A review of multiobjective test problems and a scalable test problem toolkit.
\textit{IEEE Transactions on Evolutionary Computation} \textbf{10}(5), 477--506 (2006)

\bibitem{jiang2017benchmark}
Jiang, S., Yang, S., Zhang, S., Ong, Y.-S., Yao, X.:
Benchmark problems for dynamic multiobjective optimization with changing objectives.
\textit{IEEE Transactions on Cybernetics} \textbf{47}(9), 1--14 (2017)


\bibitem{Fu2024TBM}
Fu, X., Wu, M., Tiong, R.L.K., Zhang, L.:
Data-driven joint multi-objective prediction and optimization for advanced control during tunnel construction.
\textit{Expert Systems with Applications} \textbf{238}, 122118 (2024)

\bibitem{Yang2023EvCo}
Yang, C., Ding, J., Jin, Y., Chai, T.:
A data stream ensemble assisted multifactorial evolutionary algorithm for offline data-driven dynamic optimization.
\textit{Evolutionary Computation} \textbf{31}(4), 433--458 (2023)

\bibitem{Liu2025STMOEA}
Liu, Z., Wang, H., Gong, M., and Jin, Y.:
Data stream driven dynamic multiobjective optimization using surrogate transfer.
\textit{IEEE Transactions on Emerging Topics in Computational Intelligence} (2025). Early Access.

\bibitem{Qiao2025SDRAMTCSO}
Qiao, J., Huang, W., and Meng, X.:
Data-driven dynamic multiobjective optimization with response to stochastic changes for municipal solid waste incineration process.
\textit{IEEE Transactions on Evolutionary Computation} (2025).doi:10.1109/TEVC.2025.3592956

\bibitem{Zhou2020DMOEALEM}
Zhou, Y., Wang, H., Jin, Y.:
Learning evolution model for dynamic multiobjective optimization.
\textit{IEEE Transactions on Evolutionary Computation} \textbf{24}(2), 253--267 (2020)

\bibitem{Li2020DVCDMOEA}
Li, J., Wang, H., Jin, Y., Gong, M.:
Dynamic multiobjective evolutionary algorithm with variable classification.
\textit{IEEE Transactions on Cybernetics} \textbf{50}(9), 3942--3954 (2020)

\bibitem{goldberg1987dynamic}
Goldberg, D.E., Smith, R.E.:
Nonstationary function optimization using genetic algorithms with dominance and diploidy.
In: \textit{Proceedings of the Second International Conference on Genetic Algorithms}, pp.~59--68 (1987)

\bibitem{dasgupta1992dynamic}
Dasgupta, D., McGregor, D.R.:
Nonstationary function optimization using the structured genetic algorithm.
In: \textit{Parallel Problem Solving from Nature (PPSN)}, pp.~145--154 (1992)

\bibitem{li2008gdbg}
Li, C., Yang, S.:
A generalized approach to construct benchmark problems for dynamic optimization.
In: \textit{Simulated Evolution and Learning (SEAL)}, pp.~391--400 (2008)

\bibitem{li2013gmpb}
Li, C., Yang, S., Nguyen, T.T.:
Benchmark generator for dynamic optimization problems with rotation.
\textit{IEEE Transactions on Evolutionary Computation} \textbf{17}(2), 264--278 (2013)

\bibitem{branke1999mpb}
Branke, J.: Memory enhanced evolutionary algorithms for changing optimization problems.
In: Proceedings of the Congress on Evolutionary Computation (CEC), pp. 1875--1882 (1999)

\bibitem{Zhong2025DASE}
Zhong, Y.-T., and Gong, Y.-J.:
Data-driven evolutionary computation under continuously streaming environments: A drift-aware approach.
\textit{IEEE Transactions on Evolutionary Computation} (2025). doi:10.1109/TEVC.2025.3589643

\bibitem{deb2007dynamic}
Deb, K., Rao N., U.B., Karthik, S.:
Dynamic Multi-Objective Optimization and Decision-Making Using Modified NSGA-II.
In: Simulated Evolution and Learning, pp. 329–338. Springer, LNCS 4684 (2007)

\bibitem{Jiang2018CEC}
Jiang, S., Yang, S., Yao, X., Tan, K.C., Kaiser, M., Krasnogor, N.:
Benchmark problems for the CEC2018 competition on dynamic multiobjective
optimisation. Technical Report, School of Computing,
Newcastle University (2018)

\bibitem{coello2004igd}
Coello Coello, C.A., Cort\'es, N.C.: 
A study of convergence and diversity metrics for multi-objective optimization. 
In: Evolutionary Multi-Criterion Optimization (EMO), pp. 894--908 (2004)

\bibitem{Muruganantham2016KF}
Muruganantham, A., Tan, K.C., Vadakkepat, P.:
Evolutionary dynamic multiobjective optimization via Kalman filter prediction.
\textit{IEEE Transactions on Cybernetics}
\textbf{46}(12), 2862--2873 (2016).
doi:10.1109/TCYB.2015.2490738

\bibitem{Li2024MSAS}
Li, H., Wang, Z., Lan, C., Wu, P., Zeng, N.:
A novel dynamic multiobjective optimization algorithm with non-inductive
transfer learning based on multi-strategy adaptive selection.
\textit{IEEE Transactions on Neural Networks and Learning Systems}
\textbf{35}(11), 16533--16547 (2024).
doi:10.1109/TNNLS.2023.3295461

\bibitem{Zhang2020PBDMO}
Zhang, Q., Yang, S., Jiang, S., Wang, R., Li, X.:
Novel prediction strategies for dynamic multi-objective optimization.
\textit{IEEE Transactions on Evolutionary Computation}
\textbf{24}(2), 260--274 (2020).
doi:10.1109/TEVC.2019.2922834

\end{thebibliography}
\end{document}


\maketitle

\section{Complete Drift-Pattern Definitions}
\label{sec:supp_definitions}

This supplementary material provides the complete mathematical definitions of the drift patterns, the performance metric used for quantitative evaluation, and additional temporal tracking visualizations for SDDMO-Bench.

\subsection{Drift Signal Modeling}
\label{sec:supp_drift}

For an environment index $\tau\in\{0,1,\ldots,T-1\}$, SDDMO-Bench defines a scalar drift signal as
\begin{equation}
\delta_{\tau}=d(\tau;P,\epsilon,k,T),
\label{eq:drift_general}
\end{equation}
where $P$, $\epsilon$, $k$, and $T$ denote the drift period, noise magnitude, number of abrupt changes, and total number of environments, respectively. Depending on the selected drift pattern, only a subset of these parameters is used.

Six representative patterns are considered to model the absence of additional distributional drift, irregular abrupt changes, periodic abrupt changes, smooth incremental changes, noisy incremental changes, and hybrid extreme changes.

\subsubsection{D1: No Additional Distributional Drift}

The first pattern is defined as
\begin{equation}
\delta_{\tau}=0.
\label{eq:d1}
\end{equation}
D1 disables the additional sampling transformation, while the selected FDA/DF base function may retain its intrinsic temporal dynamics. It therefore provides a reference condition for isolating the effect of the additional distributional drift.

\subsubsection{D2: Abrupt Drift}

Let $k'=\min(k,T-1)$, $\lambda_0=0$, and $\lambda_{k'+1}=T$. The jump indices $\{\lambda_1,\ldots,\lambda_{k'}\}$ are sampled uniformly without replacement from $\{1,\ldots,T-1\}$ and sorted in ascending order. The drift signal is defined as
\begin{equation}
\delta_{\tau}=u_i,
\qquad
\lambda_i\leq\tau<\lambda_{i+1},
\quad
i=0,\ldots,k',
\label{eq:d2}
\end{equation}
where $u_i\sim\operatorname{Uniform}(-1,1)$. A new state is generated at each jump index and remains unchanged until the next jump, yielding piecewise-constant environments with irregular abrupt transitions.

\subsubsection{D3: Periodic Abrupt Drift}

Let $r=\tau\bmod P$ and $P_{\mathrm{half}}=\lfloor P/2\rfloor$. The drift signal is defined as
\begin{equation}
\delta_{\tau}=
\begin{cases}
-0.5, & r<P_{\mathrm{half}},\\
\phantom{-}0.5, & r\geq P_{\mathrm{half}}.
\end{cases}
\label{eq:d3}
\end{equation}
D3 periodically alternates between two discrete states, representing recurring abrupt changes or repeated operating regimes.

\subsubsection{D4: Periodic Incremental Drift}

The periodic incremental drift is defined as
\begin{equation}
\delta_{\tau}
=
\sin\left(
\frac{2\pi\tau}{P}+\frac{\pi}{4}
\right).
\label{eq:d4}
\end{equation}
D4 represents smooth and continuous periodic variation. The phase shift avoids starting the dynamic process at a zero crossing.

\subsubsection{D5: Periodic Incremental and Noisy Drift}

D5 adds stochastic perturbations to the periodic incremental drift:
\begin{equation}
\delta_{\tau}
=
\sin\left(
\frac{2\pi\tau}{P}+\frac{\pi}{4}
\right)
+
\epsilon\xi_{\tau},
\qquad
\xi_{\tau}\sim\operatorname{Uniform}(-1,1).
\label{eq:d5}
\end{equation}
The sinusoidal component determines the periodic trend, whereas $\epsilon\xi_{\tau}$ introduces local irregular fluctuations around this trend.

\subsubsection{D6: Hybrid Extreme Drift}

D6 combines high-frequency oscillation, stochastic noise, and random abrupt jumps:
\begin{equation}
\delta_{\tau}
=
\sin\left(
\frac{4\pi\tau}{P}
\right)
+
\epsilon\xi_{\tau}
+
J_{\tau},
\label{eq:d6}
\end{equation}
where $\xi_{\tau}\sim\operatorname{Uniform}(-1,1)$ and
\begin{equation}
J_{\tau}=
\begin{cases}
\eta_{\tau}, & z_{\tau}=1,\\
0, & z_{\tau}=0,
\end{cases}
\qquad
z_{\tau}\sim\operatorname{Bernoulli}(0.1),
\quad
\eta_{\tau}\sim\operatorname{Uniform}(-2,2).
\label{eq:jump}
\end{equation}
The variable $z_{\tau}$ determines whether a jump occurs, while $\eta_{\tau}$ determines the corresponding jump magnitude.

\begin{figure}[!t]
\centering
\includegraphics[width=\linewidth]{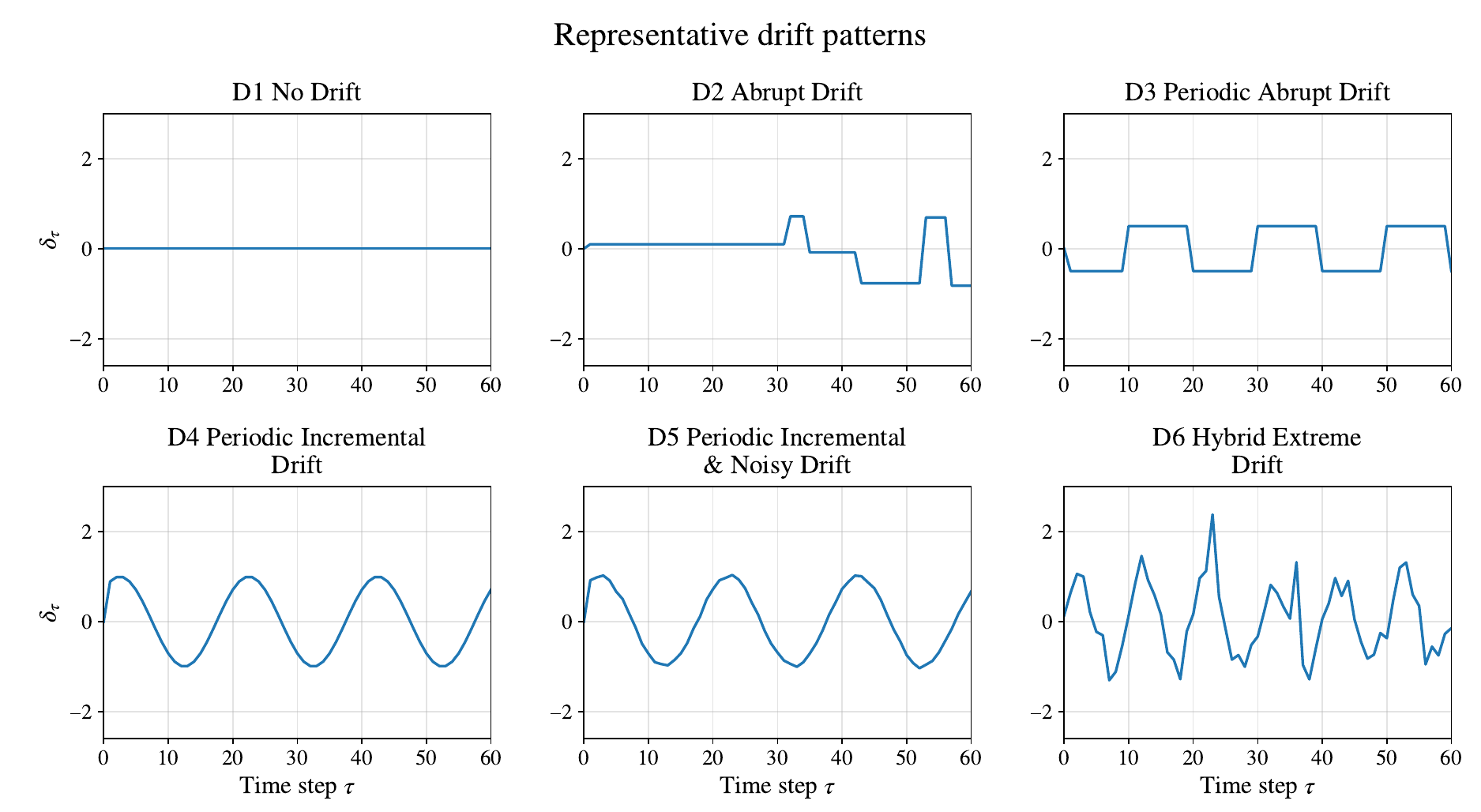}
\caption{Temporal behaviors of drift patterns D1--D6.}
\label{fig:drift_patterns}
\end{figure}

\begin{figure}[!t]
\centering
\includegraphics[width=\linewidth]{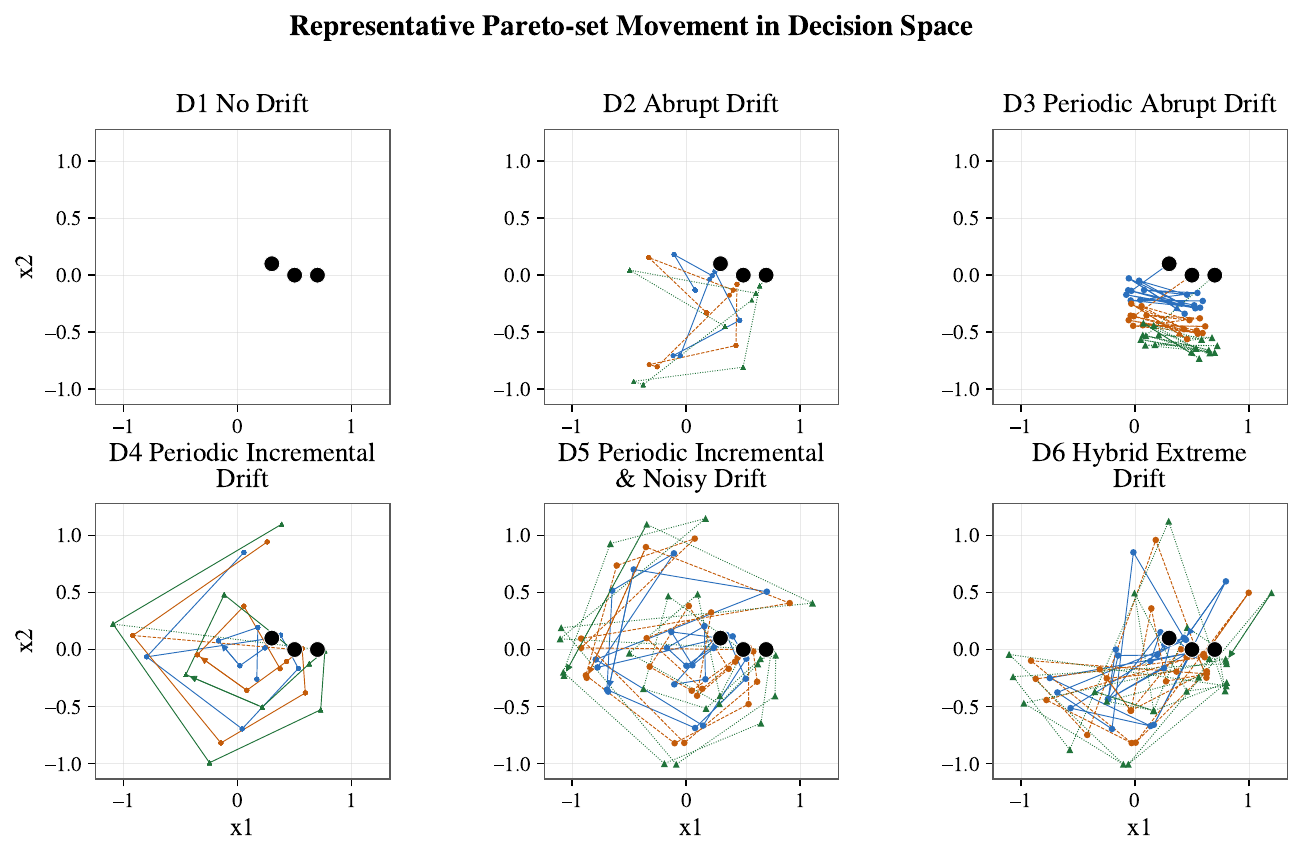}
\caption{Movement of representative transformed decision samples in a two-dimensional decision subspace under drift patterns D1--D6. Black markers denote the initial anchor samples.}
\label{fig:sample_movement}
\end{figure}

D1--D6 represent qualitatively different forms of temporal non-stationarity rather than monotonically increasing difficulty levels. The drift signal controls the distribution and coverage of the observed decision vectors through $\Phi(\cdot,\delta_{\tau})$, whereas the time-dependent base function $\mathbf{F}_{\tau}$ controls the objective mapping and the associated Pareto structures. Fig.~\ref{fig:drift_patterns} shows the six drift signals, while Fig.~\ref{fig:sample_movement} visualizes the corresponding movements of representative transformed decision samples.

\FloatBarrier

\section{Performance Metric}
\label{sec:supp_metric}

The mean inverted generational distance (MIGD) evaluates the ability of an algorithm to track the evolving Pareto fronts. Let $\mathcal{A}_{\tau}\subset\mathbb{R}^{M}$ denote the approximation front obtained in environment $\tau$, and let $PF_{\tau}^{\star}\subset\mathbb{R}^{M}$ denote the corresponding reference Pareto front. The environment-wise IGD and its temporal average are defined as
\begin{equation}
\begin{aligned}
\operatorname{IGD}_{\tau}
&=
\frac{1}{|PF_{\tau}^{\star}|}
\sum_{\mathbf{y}\in PF_{\tau}^{\star}}
\min_{\mathbf{a}\in\mathcal{A}_{\tau}}
\left\|
\mathbf{a}-\mathbf{y}
\right\|_2,\\
\operatorname{MIGD}
&=
\frac{1}{T}
\sum_{\tau=0}^{T-1}
\operatorname{IGD}_{\tau}.
\end{aligned}
\label{eq:igd_migd}
\end{equation}
A lower MIGD indicates better convergence, coverage, and temporal tracking performance. For each environment, $PF_{\tau}^{\star}$ is generated by evaluating the time-dependent base function $\mathbf{F}_{\tau}$ over the complete feasible domain. Because $\Phi(\cdot,\delta_{\tau})$ controls only the distribution of the observed decision samples, it is not applied when constructing the reference Pareto-optimal set or Pareto front. Consequently, the metric evaluates whether an algorithm can recover the underlying Pareto structure from the temporally changing and distributionally shifted observations.

\FloatBarrier

\section{Temporal Tracking Visualizations}
\label{sec:supp_visualizations}

\subsection{Visualization Setup}
\label{sec:supp_visualization_setup}

To complement the aggregate MIGD results, this section provides environment-level visualizations of the approximation fronts obtained by six representative algorithms on F3 (DF10). F3 is selected as a representative benchmark instance to illustrate environment-specific differences in convergence, coverage, solution dispersion, and post-change recovery.

For each algorithm, the six rows correspond to drift patterns D1--D6, while the three columns represent the initial, intermediate, and final environments, i.e., $\tau=0$, $\tau=30$, and $\tau=59$. Blue markers denote the approximation solutions returned by the algorithm, whereas brown surfaces denote the reference Pareto fronts generated from the corresponding time-dependent base function.

\subsection{Qualitative Observations}
\label{sec:supp_qualitative_observations}

Figs.~\ref{fig:dnsga_f3}--\ref{fig:tbm_f3} show that algorithmic differences become more apparent after the initial environment. Under D1, the deviations mainly reflect the ability to track the intrinsic dynamics of DF10. D2 and D3 reveal differences in adaptation to abrupt transitions, whereas D4 and D5 emphasize tracking under incremental changes, with D5 additionally introducing stochastic fluctuations. D6 combines oscillation, noise, and abrupt jumps, resulting in more heterogeneous tracking behavior.

These visualizations complement MIGD by indicating whether tracking errors arise from insufficient convergence, incomplete coverage, uneven dispersion, or delayed recovery. Since only three representative environments are displayed, they provide qualitative evidence rather than an independent algorithm ranking; the quantitative comparison remains based on MIGD over all $T=60$ environments.

\subsection{Algorithm-Wise Results}
\label{sec:supp_algorithm_results}

\begin{figure*}[!t]
\centering
\includegraphics[
    width=\textwidth,
    height=0.82\textheight,
    keepaspectratio
]{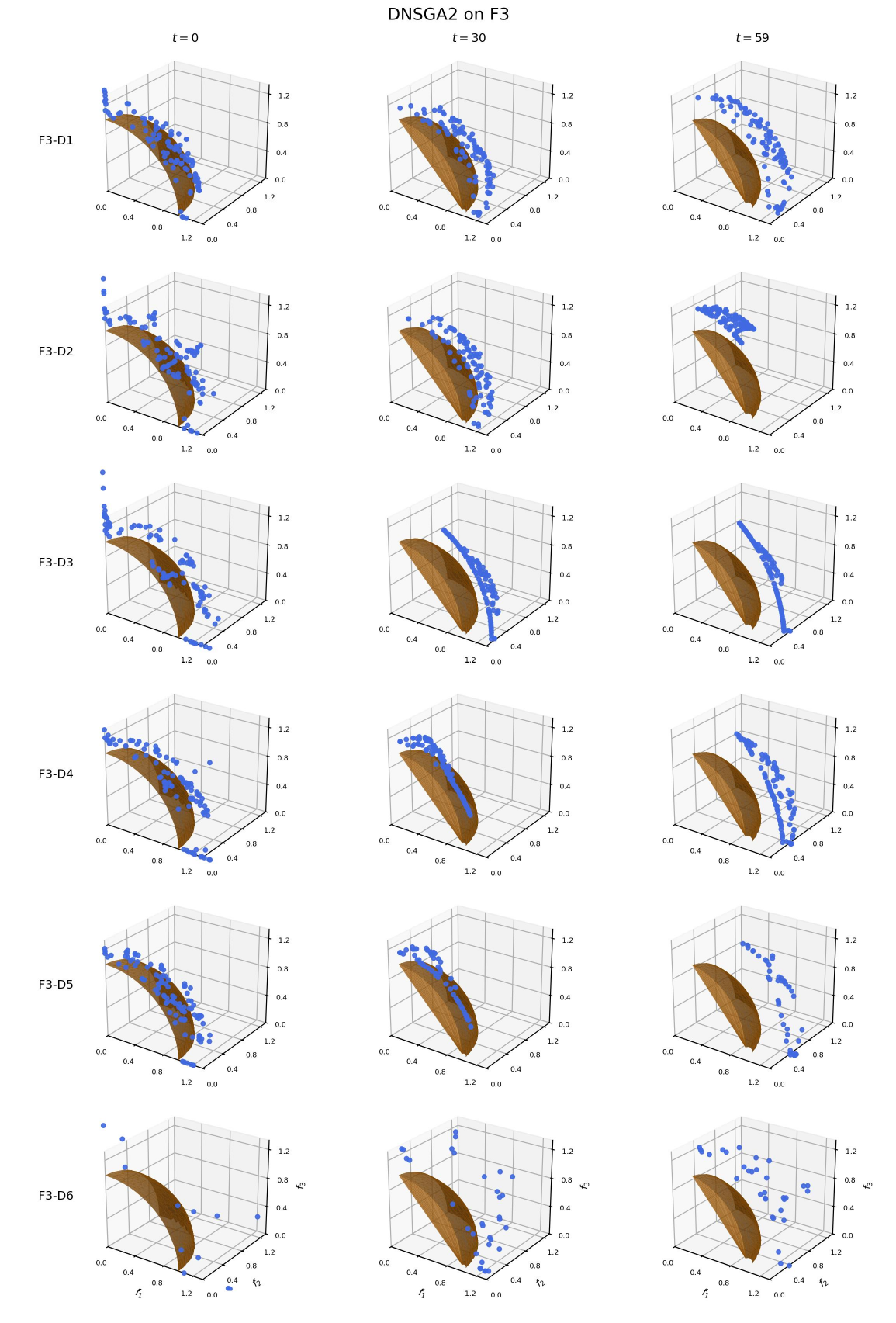}
\caption{Temporal approximation fronts obtained by DNSGA-II on F3 under drift patterns D1--D6. Rows correspond to D1--D6, and columns correspond to environments $\tau=0$, $\tau=30$, and $\tau=59$.}
\label{fig:dnsga_f3}
\end{figure*}

\begin{figure*}[!t]
\centering
\includegraphics[
    width=\textwidth,
    height=0.82\textheight,
    keepaspectratio
]{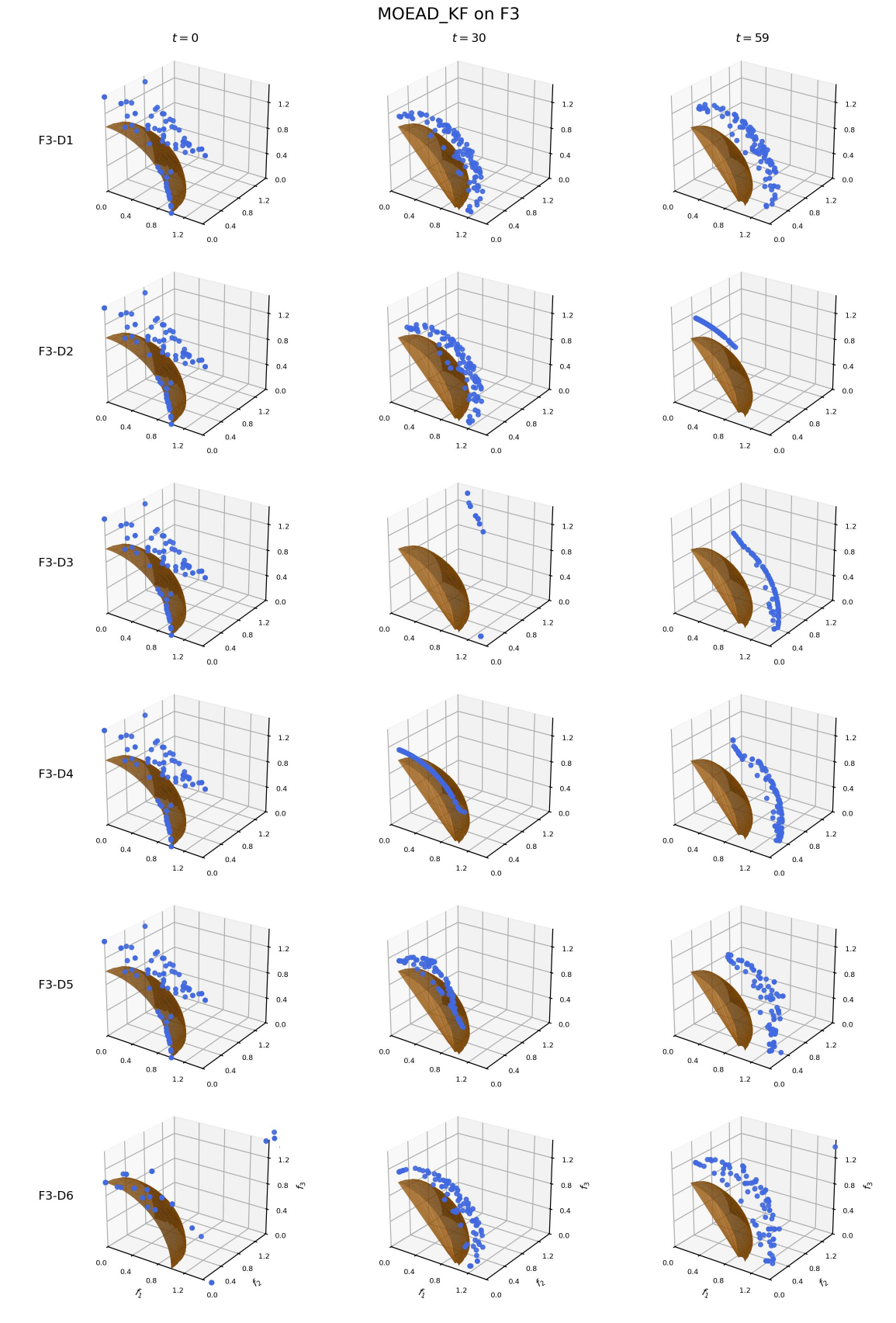}
\caption{Temporal approximation fronts obtained by MOEA/D-KF on F3 under drift patterns D1--D6. Rows correspond to D1--D6, and columns correspond to environments $\tau=0$, $\tau=30$, and $\tau=59$.}
\label{fig:moead_kf_f3}
\end{figure*}

\begin{figure*}[!t]
\centering
\includegraphics[
    width=\textwidth,
    height=0.82\textheight,
    keepaspectratio
]{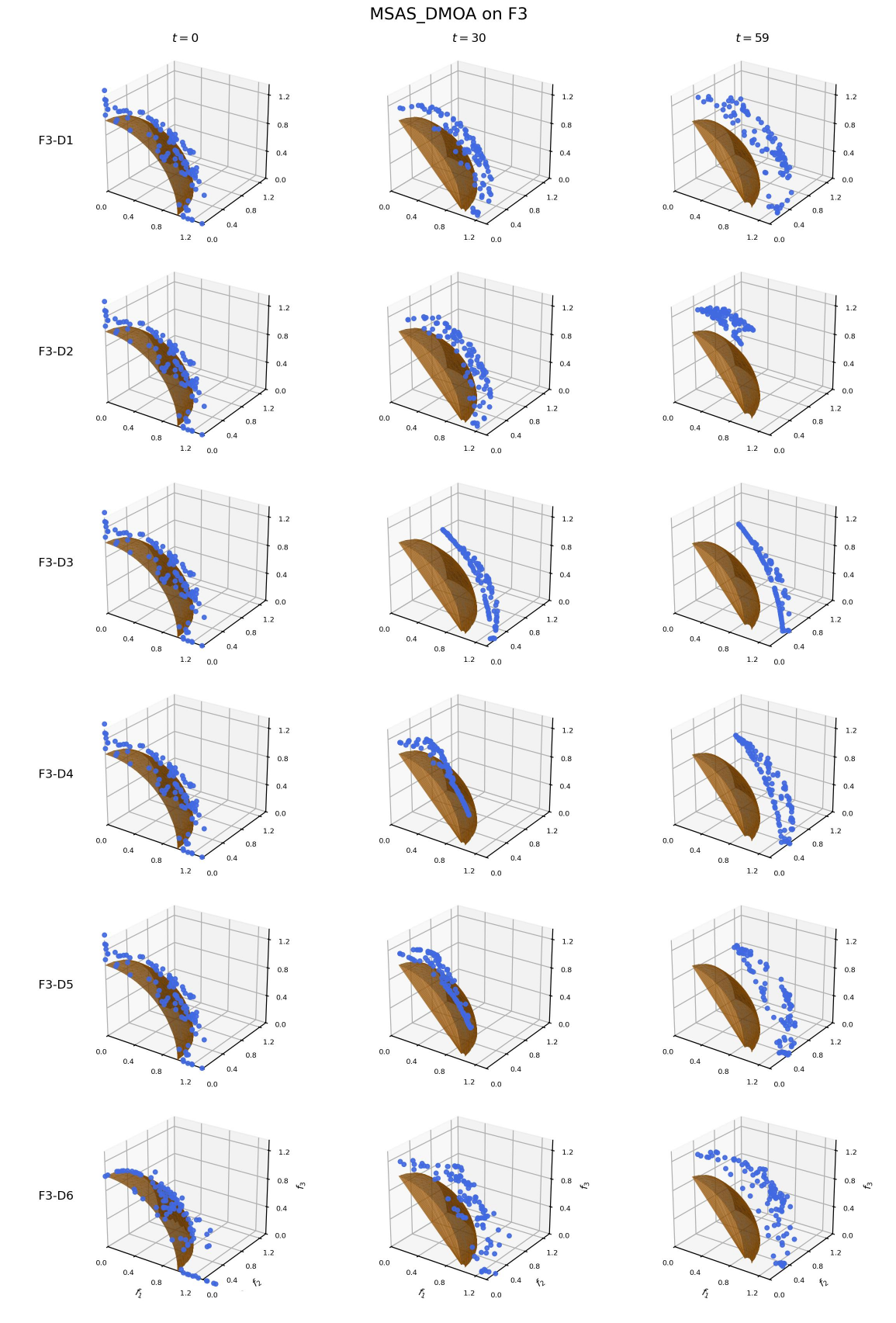}
\caption{Temporal approximation fronts obtained by MSAS-DMOA on F3 under drift patterns D1--D6. Rows correspond to D1--D6, and columns correspond to environments $\tau=0$, $\tau=30$, and $\tau=59$.}
\label{fig:msas_dmoa_f3}
\end{figure*}

\begin{figure*}[!t]
\centering
\includegraphics[
    width=\textwidth,
    height=0.82\textheight,
    keepaspectratio
]{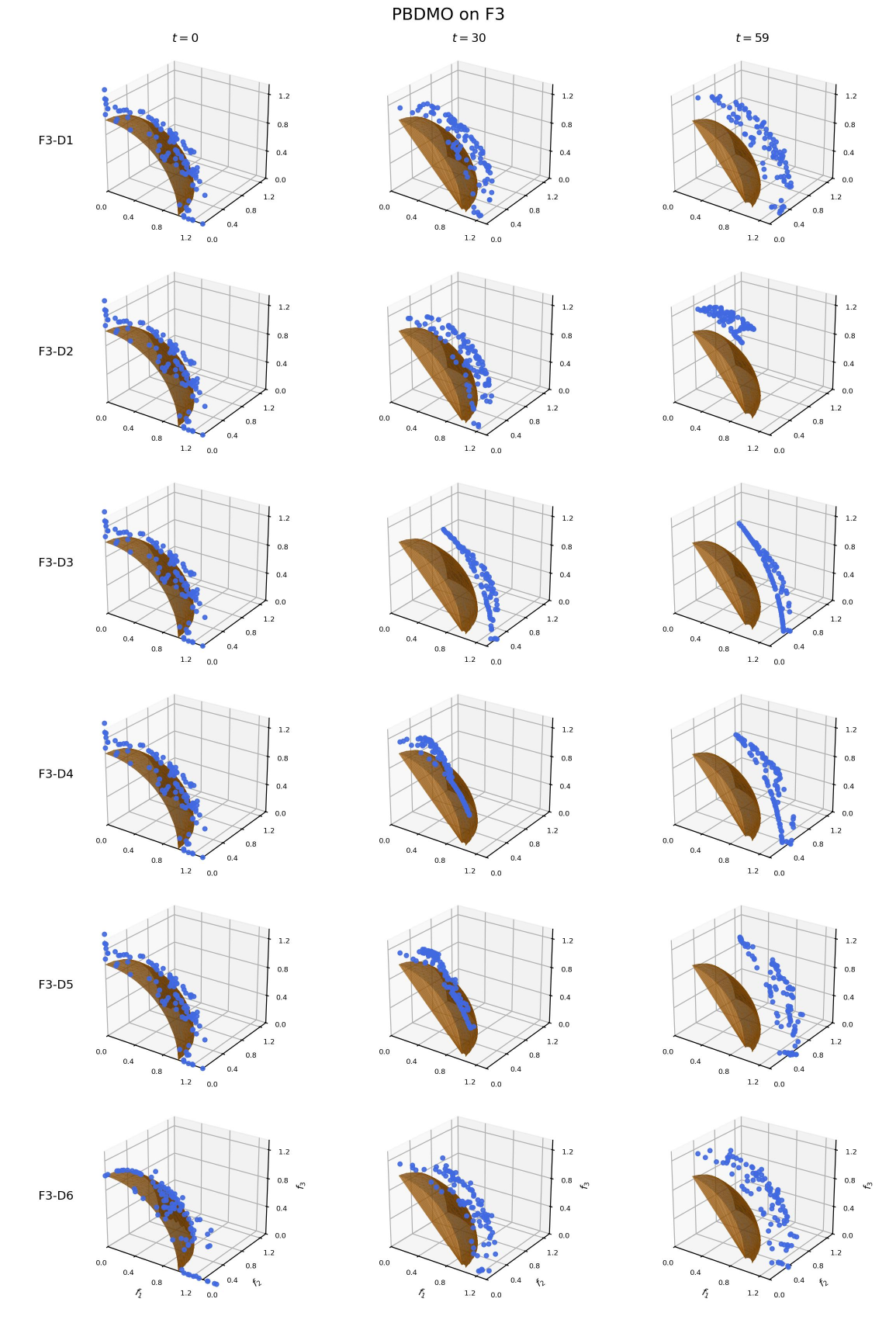}
\caption{Temporal approximation fronts obtained by PBDMO on F3 under drift patterns D1--D6. Rows correspond to D1--D6, and columns correspond to environments $\tau=0$, $\tau=30$, and $\tau=59$.}
\label{fig:pbdmo_f3}
\end{figure*}

\begin{figure*}[!t]
\centering
\includegraphics[
    width=\textwidth,
    height=0.82\textheight,
    keepaspectratio
]{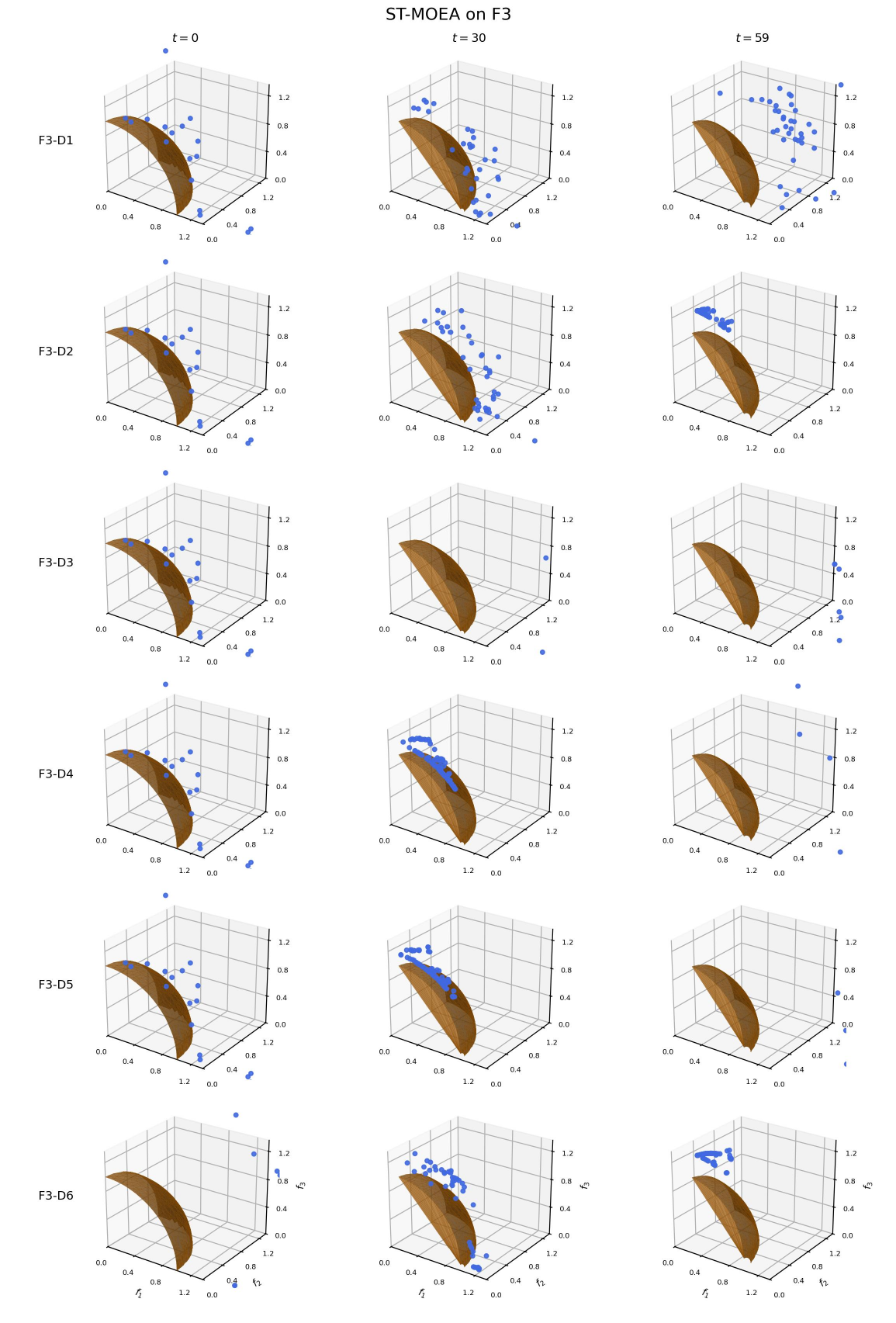}
\caption{Temporal approximation fronts obtained by ST-MOEA on F3 under drift patterns D1--D6. Rows correspond to D1--D6, and columns correspond to environments $\tau=0$, $\tau=30$, and $\tau=59$.}
\label{fig:st_moea_f3}
\end{figure*}

\begin{figure*}[!t]
\centering
\includegraphics[
    width=\textwidth,
    height=0.82\textheight,
    keepaspectratio
]{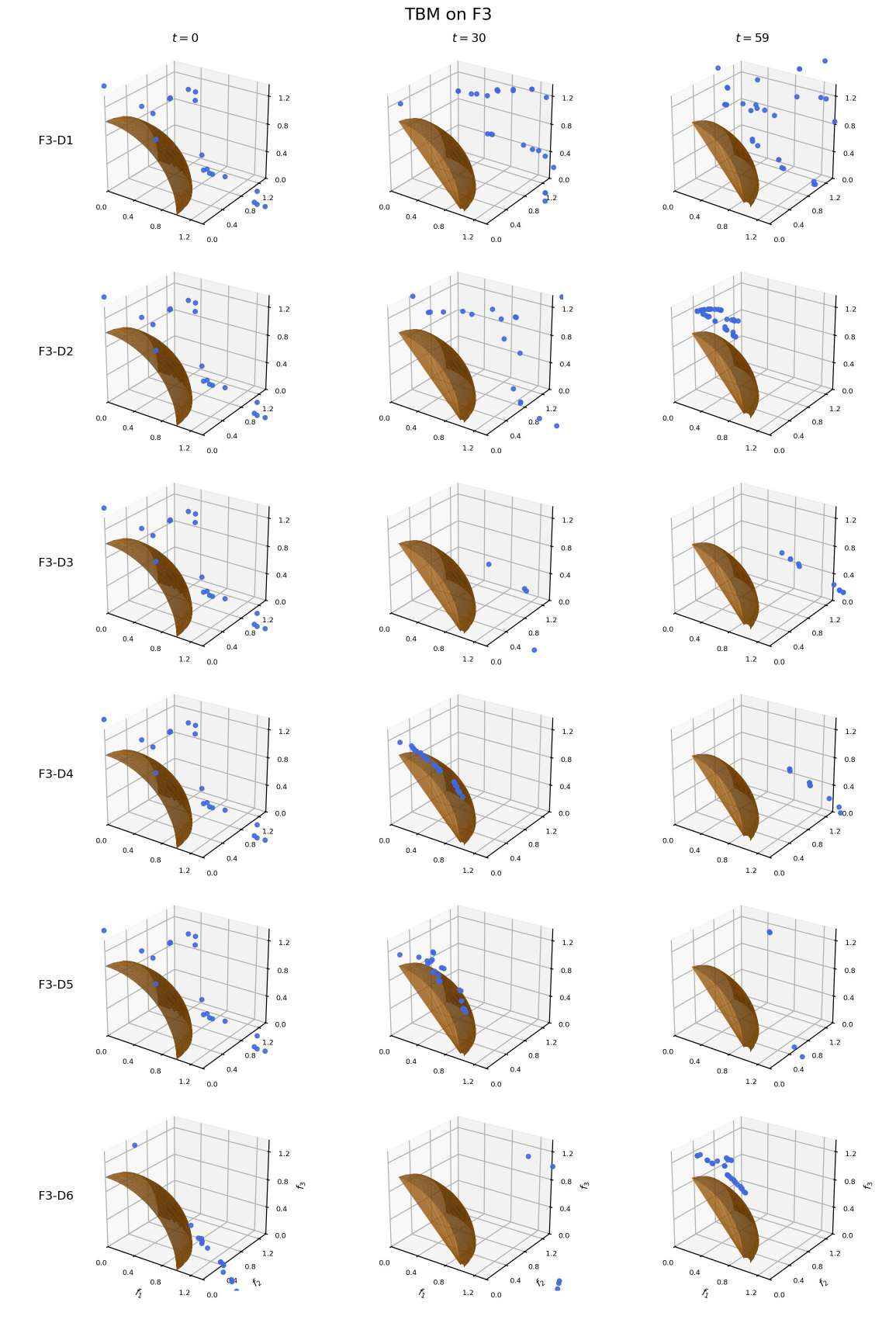}
\caption{Temporal approximation fronts obtained by TBM on F3 under drift patterns D1--D6. Rows correspond to D1--D6, and columns correspond to environments $\tau=0$, $\tau=30$, and $\tau=59$.}
\label{fig:tbm_f3}
\end{figure*}

Representative temporal snapshots of the approximation fronts obtained by DNSGA-II, MOEA/D-KF, MSAS-DMOA, PBDMO, ST-MOEA, and TBM are presented in Figs.~\ref{fig:dnsga_f3}--\ref{fig:tbm_f3}, respectively. All figures use the same selected environments, reference fronts, viewpoint, and visual encoding to support direct comparisons across algorithms and drift patterns.

\FloatBarrier